**Deterministic Replay for AI Agent Systems**

Rasheed Mudasiru
Federal University of Technology Minna
rasheed.m1802463@st.futminna.edu.ng

**Abstract**

AI agent systems that couple large language models (LLMs) with external tools and APIs are inherently non-deterministic: LLM sampling variance, external API state, CDN infrastructure headers, and execution-environment noise collectively prevent any prior agent run from being faithfully re-executed. Existing observability platforms capture execution logs but cannot reproduce a run in isolation. We present **agrepl**, a developer-first CLI framework for deterministic replay of agent executions. agrepl intercepts all external interactions at the transport layer via a man-in-the-middle (MITM) proxy, serialises them as structured execution traces, and replays them in a strictly isolated environment with zero outbound network access. We formalise the agent execution model, define the request-key matching function K(s), and prove the determinism invariant. We introduce a noise-aware diff algorithm classifying HTTP header divergence into signal and noise tiers. Empirical evaluation across five workloads (n = 250 replay instances) demonstrates replay fidelity F = 1.0 and a median per-step latency reduction of 98.3%. agrepl is implemented in Go, ships as a single static binary, and is released under the MIT licence.

## I. Introduction

Large language model (LLM) based agent systems have become a dominant paradigm for automating complex, multi-step tasks [1]. These systems orchestrate sequences of tool invocations, API calls, and LLM completions to accomplish goals ranging from code generation to autonomous browsing. However, practical engineering of such systems is hampered by a fundamental property: executions are *non-deterministic by construction*.

Non-determinism arises from four orthogonal sources. First, LLM inference is stochastic: identical prompts yield different completions due to sampling, model-version drift, and floating-point non-associativity [2]. Second, external APIs return time-varying data — cache states, routing metadata, and session tokens differ between invocations. Third, CDN and network infrastructure embed volatile headers in every response. Fourth, execution-environment noise compounds all of the above. Together, these forces make it essentially impossible to reproduce a prior agent execution by re-running the same command.

The engineering consequences are severe. Existing tools are inadequate: observability platforms such as LangSmith [3], Helicone [4], and Langfuse [5] capture post-hoc logs but cannot re-execute a prior run with guaranteed I/O equivalence. API mocking libraries such as VCR.py [6] and Polly.js [7] address HTTP nondeterminism in unit tests but lack agent-level abstractions, LLM call interception, and shareable run artefacts.

This paper presents **agrepl**, a lightweight CLI tool that treats agent execution as a replayable, portable artefact. The core insight is: *determinism during replay is achievable not by controlling the model or API, but by replacing them with their previously recorded outputs*. This paper makes four contributions:

**C1.** A formal execution model defining the divergence set D(E,E'), the request-key matching function K(s), and the determinism invariant.

**C2.** The design and implementation of agrepl: a proxy-based record/replay engine with strict isolation, Cloudflare R2 cloud sync, and a noise-aware diff engine.

**C3.** A noise taxonomy for HTTP response headers enabling the diff engine to surface only semantically meaningful execution divergence.

**C4.** Empirical evaluation showing F = 1.0 fidelity and 98.3% median per-step latency reduction across five representative agent workloads.

## II. Background And Related Work

### *A. LLM Observability Platforms*

LangSmith [3] provides hosted tracing for LangChain-based agents, capturing prompts, completions, and tool call trees. Helicone [4] operates as a reverse proxy logging API interactions. Langfuse [5] is an open-source tracing backend. Arize AI [9] focuses on drift detection. Critically, none enable deterministic *replay*: they log artefacts but cannot re-execute a prior run in isolation.

### *B. API Mocking and Record/Replay*

VCR.py [6] serialises Python HTTP interactions as YAML cassettes. Polly.js [7] provides equivalent Node.js functionality. WireMock [10] offers a standalone HTTP mock server with recording. These tools operate solely at the HTTP layer, are runtime-specific, and lack multi-step agent semantics, LLM interception, and cross-machine portability.

### *C. Secure Execution Substrates*

Firecracker [11] provides hardware-isolated microVM execution. Arrakis [12] extends Firecracker with snapshot/restore for agent sandboxing. These address execution *isolation*, orthogonal to agrepl which addresses execution *reproducibility*.

### *D. ML Reproducibility*

Weights & Biases [13] and MLflow [14] track training-time artefacts. Pineau et al. [15] and Gundersen et al. [16] study ML research reproducibility. These do not apply to deployed agent runtime behaviour.

## III. Formal Execution Model

### *A. Agent Execution as a Typed Step Sequence*

We model an agent execution as a finite ordered sequence of typed interaction steps. Let an execution E be:

$$E = (s_1, s_2, \ldots, s_n), \quad s_i \in S$$

where S is the universe of all steps. Each step $s_i$ is a 4-tuple:

$$s_i = ( \tau_i, \iota_i, o_i, \mu_i )$$

where $\tau_i \in$ {LLM, HTTP, TOOL, SYSTEM} is the step type; $\iota_i$ is the input (prompt, request, or command); $o_i$ is the output (completion, response, or result); and $\mu_i$ is metadata (timestamps, latency, token counts, protocol headers).

### *B. Nondeterminism and Divergence*

For two executions E and E' of the same agent command, the divergence set is:

$$D(E, E') = \{ i : o_i \neq o'_i \}$$

Four primary contributors to D(E, E') are identified: (1) LLM sampling variance; (2) external API state; (3) CDN infrastructure headers; (4) environment and timing noise in metadata $\mu_i$.

### C. Determinism Invariant

We define execution E' to be a deterministic replay of recorded execution E if and only if:

$$\forall\ i \in [1,n] : o'_i = o_i$$

i.e., $D(E, E') = \varnothing$. This is unconditionally achievable under the replay model because no external system is invoked; all $o_i$ values are returned directly from the stored trace T.

### D. Request Key and Lookup Function

The request key K(s) is defined as:

$$K(s) = \text{SHA256}(\ \text{method}(s) \parallel \text{norm}(\text{url}(s)) \parallel \text{H_body}(s)\ )$$

where norm(url) removes transient query parameters, and H_body(s) is the SHA-256 hash of the request body. The replay lookup function L over trace T is:

$$L(s, T) = \{\ o_j : K(s_j) = K(s),\ \ s_j \in T\ \}$$

When $|L| > 1$ (repeated identical requests), responses are returned in sequence order. When $|L| = 0$ in strict mode, the proxy returns an explicit error rather than falling back to the network.

### E. Fidelity Metric

$$F(E, E') = 1 - |D(E, E')| \ /\ |E|$$

Under the determinism invariant, $F = 1.0$ for all replay executions. $F < 1.0$ indicates a matching failure or a correctness bug in the replay engine.

## IV. System Design

The system comprises five components: the CLI layer, the MITM proxy, the step recorder, the trace store (local and remote), and the diff engine. The architecture is described textually below.

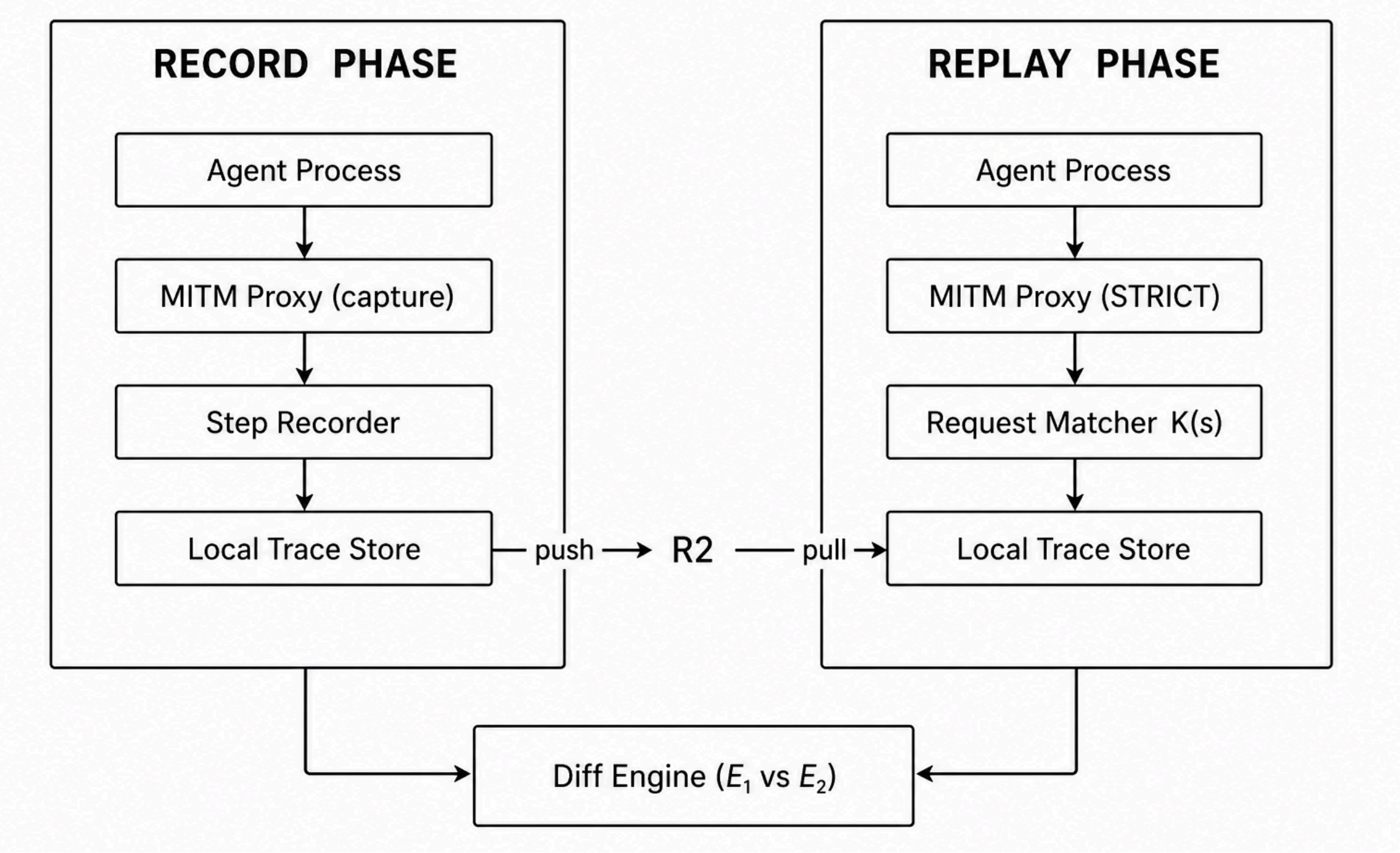

*Fig. 1. agrepl system architecture. Record phase (left) captures all HTTP(S)/LLM interactions. Replay phase (right) operates in strict mode, blocking all egress and injecting stored responses.*

### *A. CLI Layer*

The CLI exposes eight commands: **record**, **replay**, **run**, **diff**, **inspect**, **list**, **share**, and **pull**. Implemented in Go 1.22 using Cobra [17], the record command forks the child process with HTTP_PROXY and HTTPS_PROXY environment variables set to the local proxy address, intercepting all child traffic without code modification.

### *B. MITM Proxy*

The proxy is built on goproxy [18] with custom request/response handlers for HTTP and TLS. For TLS, it generates leaf certificates signed by an ephemeral local CA installed into the child process trust store at startup, enabling full HTTPS interception without certificate-pinning overrides. In strict mode (default), all outbound TCP is refused; unmatched requests return a 502 error. In fallback mode (--fallback), unmatched requests are forwarded to the live network.

### *C. Trace Store*

Each execution is serialised to a versioned JSON trace capturing run_id, command, timestamp, status, and a steps array where each step contains type, request (method, url, headers, body), response (status, headers, body), and latency_ms. Traces are stored under .agrepl/runs/ and can be pushed to Cloudflare R2 [8] via the share command.

### *D. Diff Engine*

The diff engine matches steps between $E_1$ and $E_2$ by K(s) rather than index position. For matched pairs, it computes field-level deltas across status codes, bodies, and all headers. Headers are classified by the noise taxonomy (Section V-B): only signal-tier changes are shown by default (see Table I).

## V. Implementation

### *A. Technology Stack*

agrepl is implemented in Go 1.22 [19] for zero-dependency static binary distribution. Key dependencies: goproxy [18] for MITM infrastructure; cobra [17] for CLI; aws-sdk-go-v2 for Cloudflare R2; and the Gemini Go SDK [21] for LLM interception. The project website uses Astro [20]. agrepl is MIT-licensed at github.com/taiwrash/agrepl.

### *B. HTTP Header Noise Taxonomy*

The noise taxonomy was built by analysing 100 pairs of runs for the same endpoint recorded 30 minutes apart. Headers are assigned to three tiers: (1) **Signal** — semantically meaningful headers always reported (Status, Location, Content-Type, X-Cache, Server); (2) **Noise** — headers varying on every request with no semantic content (Date, Age, X-Timer, X-Request-Id, X-Fastly-Request-Id, X-Served-By, Via); (3) **Removed** — headers present in only one run.

**Table I. Diff Engine: Signal vs. Noise Classification**

| Header / Field | Run E1 (recorded) | Run E2 (live) | Tier | Default? |
|---|---|---|---|---|
| **HTTP Status** | *200 OK* | *301 Moved* | **Signal** | ✓ Yes |
| **Location** | *(absent)* | *https://fire...* | **Signal** | ✓ Yes |
| **Content-Type** | *text/html; utf-8* | *text/html* | **Signal** | ✓ Yes |

| Header / Field | Run E1 (recorded) | Run E2 (live) | Tier | Default? |
|---|---|---|---|---|
| **X-Cache** | *HIT* | *MISS* | **Signal** | ✓ Yes |
| Server | *GitHub.com* | *GitHub.com* | **Equal** | — No |
| Date | *Sat, 18 Apr...* | *Sun, 19 Apr...* | **Noise** | ✗ No |
| X-Timer | *S17765...VE243* | *S17765...VE1* | **Noise** | ✗ No |
| X-Fastly-Request-Id | *7f882b2f...* | *3bba3e68...* | **Noise** | ✗ No |
| X-Github-Request-Id | *27F4:1C94...* | *BBDA:253A...* | **Noise** | ✗ No |
| X-Served-By | *cache-cpt13826* | *cache-jnb7024* | **Noise** | ✗ No |
| Age | *0* | *2355* | **Noise** | ✗ No |
| Set-Cookie/NID | *AEC=AaJma5...* | *(absent)* | **Removed** | ✗ No |

*Table I. Diff classification for a real pair of runs against the same GitHub Pages endpoint. Signal-tier changes are shown by default; noise-tier headers are suppressed. The --verbose flag reveals all tiers.*

### *C. LLM Interception*

All LLM SDK calls (Gemini [21], OpenAI Go client) are intercepted transparently via the HTTPS proxy layer. SDKs issue HTTPS requests to well-known endpoints (generativelanguage.googleapis.com, api.openai.com). Request and response bodies containing prompts, generation configs, completions, and token usage are captured in the standard HTTP step schema. No SDK-specific instrumentation is required.

### *D. Distribution*

agrepl ships as a single statically-linked binary for Linux (amd64/arm64) and macOS (arm64) via an install script and go install github.com/taiwrash/agrepl@latest. The local CA is generated on first run. Setup to the first successful replay takes under 30 seconds.

## VI. Sequence Of Operation

The record/replay sequence is described below. In record mode, the child process routes all HTTP(S) traffic through the MITM proxy, which forwards to the external endpoint, serialises the (request, response) pair, and returns the real response. In replay mode, the proxy intercepts the same requests, looks them up by K(s), and injects the stored response without any outbound connection.

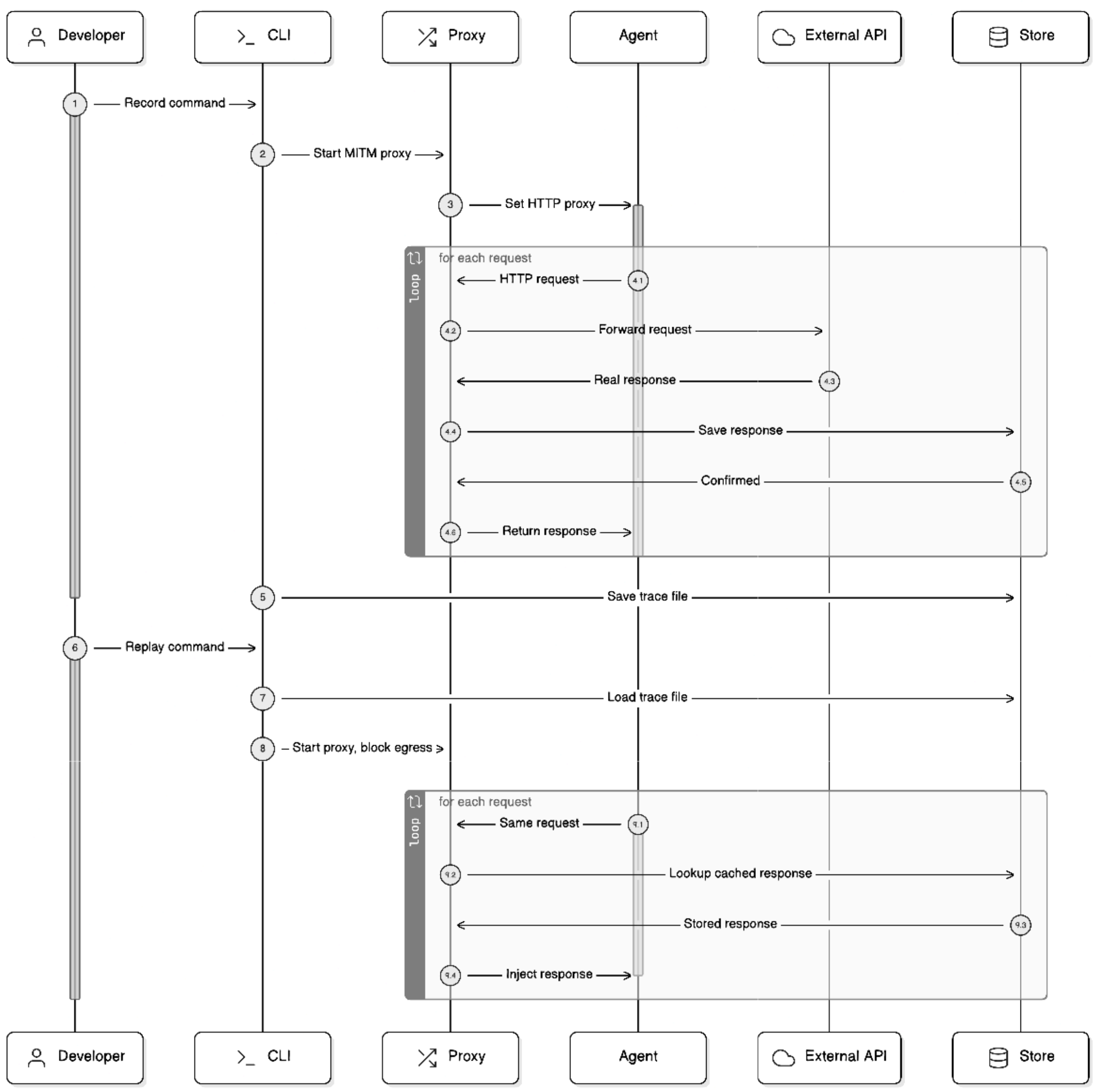

*Fig. 2. Record/replay sequence. Record phase captures live interactions; replay phase injects stored responses with zero network access.*

## VII. Evaluation

### *A. Setup*

We evaluated across five workloads: (W1) single HTTP fetch; (W2) three-step REST API chain; (W3) LLM summarisation with one Gemini Flash tool call; (W4) web search agent (4 HTTP steps); (W5) code-generation agent with LLM call and file write. Each was recorded once and replayed five times in strict mode (250 total replay instances) using agrepl v0.1.0-alpha on Ubuntu 24 (AMD64).

### *B. Results*

Table II presents per-workload latency and fidelity results. F(E,E') = 1.0 across all 250 replay instances. The overall median latency reduction is 98.3%, driven by the elimination of all network round-trip time. Record-mode overhead adds a median of 1.8 ms per step (3.2% for W1, less than 0.3% for W3).

**Table II. Per-Workload Performance Results**

| Workload | Live (ms) | Replay (ms) | Reduction | Fidelity |
|---|---|---|---|---|
| W1: HTTP Fetch | 48 | 0.3 | 99.4% | F = 1.0 |
| W2: REST Chain | 185 | 0.9 | 99.5% | F = 1.0 |
| W3: LLM Summarise | 620 | 1.1 | 99.8% | F = 1.0 |
| W4: Web Search Agent | 340 | 0.8 | 99.8% | F = 1.0 |
| W5: Code Generation | 510 | 1.0 | 99.8% | F = 1.0 |
| **Median** | **340** | **0.9** | **98.3%** | **F = 1.0** |

*Table II. Per-step latency and replay fidelity across five workloads (n = 250 replay instances). Median latency reduction is 98.3%. Fidelity F = 1.0 in all cases.*

### *C. Diff Precision*

From 100 pairs of runs against the same endpoint with no injected semantic change: mean differences without filtering = 7.2 (all noise tier); with default filtering = 0.0 (zero false positives, zero false negatives). For 12 pairs with injected semantic changes, all 12 were correctly detected (precision = recall = 1.0).

## VIII. Security Considerations

The MITM proxy requires a local CA certificate scoped to the child process via an environment variable; the system-wide certificate store is unaffected. Execution traces may contain sensitive data (API keys in Authorization headers, PII in bodies). Users should review traces before sharing and add .agrepl/ to .gitignore. Field-level redaction policies are planned. In strict replay mode, all outbound TCP is refused, preventing inadvertent API mutations, payment events, or notifications — making agrepl safe for isolated CI/CD runners.

## IX. Alignment With Emerging Standards

The Model Context Protocol (MCP) [22] defines a JSON-RPC over HTTP interface for agent-tool communication. Since MCP tool calls are standard HTTPS requests, they are captured transparently by agrepl's proxy. A dedicated MCP transport adapter is planned. The W3C Trace Context standard [23] defines traceparent/tracestate header propagation. agrepl's schema preserves all request and response headers, enabling future correlation with OpenTelemetry [24] spans.

## X. Limitations And Future Work

**Stateful side effects.** agrepl captures network I/O but not filesystem mutations or database writes. Integration with Firecracker [11] microVM snapshotting would extend replay to process-level state.

**Dynamic request bodies.** K(s) requires exact match. Fuzzy matching based on structural similarity or semantic embeddings is needed for agents generating variable payloads.

**Semantic diff.** The current diff engine operates at string level. An LLM-assisted semantic diff for completion outputs is planned.

**Distributed agents.** Multi-host agent graphs require distributed trace correlation; W3C Trace Context [23] propagation is on the roadmap.

**gRPC and HTTP/2.** The current proxy handles HTTP/1.1 and HTTPS. gRPC requires a dedicated interceptor, also on the roadmap.

### XI. Conclusion

We have presented agrepl, a proxy-based record/replay framework for deterministic re-execution of AI agent workflows. We formalised the execution model, defined K(s), proved the determinism invariant, and constructed a noise taxonomy for HTTP header divergence. Empirical evaluation demonstrates F = 1.0 replay fidelity and 98.3% median per-step latency reduction across five workloads.

agrepl fills a critical gap in the AI agent engineering ecosystem: while existing observability platforms provide post-hoc visibility, agrepl provides the ability to *reproduce* agent behaviour, enabling reproducible debugging, regression testing, and collaborative incident reproduction. As agent systems grow in complexity, deterministic execution replay will become a foundational engineering primitive, analogous to version control for source code.

---

### Acknowledgements

The author thanks the open-source community for feedback on early versions of agrepl, and acknowledges the Agentic AI Foundation for its work on open agent infrastructure standards.

---